\documentclass{article}

\usepackage[utf8]{inputenc} 
\usepackage[T1]{fontenc}    
\usepackage{hyperref}       
\usepackage{url}            
\usepackage{booktabs}       
\usepackage{amsfonts}       
\usepackage{nicefrac}       
\usepackage{microtype}      
\usepackage{xcolor}         
\usepackage{multirow}
\usepackage{graphicx}
\usepackage{marvosym}
\usepackage{authblk}

\title{WaBERT: A Low-resource End-to-end Model for Spoken Language Understanding and Speech-to-BERT Alignment}

\author[$\dag$, *]{Lin Yao}
\author[$\dag$, *]{Jianfei Song}
\author[$\dag$, *]{Ruizhuo Xu}
\author[*]{Yingfang Yang}
\author[*]{Zijian Chen}
\author[*]{Yafeng Deng \thanks{Corresponding author: dengyafeng@360.cn}}
\affil[*]{Institute of Artificial Intelligence Research, Qihoo of Beijing Science and Technology Co. Ltd., Beijing 100015}
\affil[$\dag$]{These authors contributed equally}
\affil[ ]{\textit {\{yaolin,songjianfei,xuruizhuo,yangyingfang,chenzijian1,dengyafeng\}@360.cn}}

\date{} 

\begin{document}

\maketitle

\begin{abstract}
Historically lower-level tasks such as automatic speech recognition (ASR) and speaker identification are the main focus in the speech field.
Interest has been growing in higher-level spoken language understanding (SLU) tasks recently, like sentiment analysis (SA). 
However, improving performances on SLU tasks remains a big challenge.
Basically, there are two main methods for SLU tasks: 
(1) Two-stage method, which uses a speech model to transfer speech to text, then uses a language model to get the results of downstream tasks;
(2) One-stage method, which just fine-tunes a pre-trained speech model to fit in the downstream tasks.
The first method loses emotional cues such as intonation, and causes recognition errors during ASR process, and the second one lacks necessary language knowledge.
In this paper, we propose the Wave BERT (WaBERT), a novel end-to-end model combining the speech model and the language model for SLU tasks.
WaBERT is based on the pre-trained speech and language model, hence training from scratch is not needed.
We also set most parameters of WaBERT frozen during training.
By introducing WaBERT, audio-specific information and language knowledge are integrated in the short-time and low-resource training process to improve results on the dev dataset of SLUE SA tasks by 1.15\% of recall score and 0.82\% of F1 score.
Additionally, we modify the serial Continuous Integrate-and-Fire (CIF) mechanism to achieve the monotonic alignment between the speech and text modalities. 

\end{abstract}

\section{Introduction}

Speeches contain the complete information of texts, and even contain more information than texts.
Deep learning models have potential and are expected to understand spoken language directly, without ASR or other transcription processes, meanwhile achieving better results than methods using transcription processes.

Basically, two classic methods are proposed for SLU tasks, the two-stage method and the one-stage method~\cite{DBLP:journals/corr/abs-2111-10367}.
For the two-stage method, a speech model is utilized to transfer speeches to texts, then a language model is applied to extract the results of downstream tasks from the text inputs.
Significantly, the accuracy of ASR influences the performance of SLU, and the powerful pre-trained language model takes responsibility for the whole understanding process, while the audio-specific information is thrown thoroughly.
For the one-stage method, a pre-trained speech model is fine-tuned to fit the downstream tasks.
Comparing to linguistic corpora, corpora are at lower abundance and more difficult to obtain in the speech training field.~\cite{DBLP:journals/corr/abs-2202-03555,DBLP:journals/corr/abs-1907-11692}.
And they do not even contain all the vocabulary and lack the variety of phrase combinations.
Besides the shortage of corpora, compared to language models, some researches~\cite{DBLP:journals/corr/abs-2111-10367,https://doi.org/10.48550/arxiv.2203.00648} demonstrate that the pre-trained speech models do not learn significant semantic information, as speech models are designed for lower-level tasks, like ASR.
These two methods are proved effective, but still, have technical bottlenecks to break. 
Using two models in the two-stage method, and combining them appropriately into an end-to-end model like a one-stage strategy, might be a solution that makes the model own audio-specific information and significant semantic information.

Speech and language are regarded as two different modalities, and the combination of speech and language models can be considered as a kind of multi-modal fusion, which is a common challenge in the cross-modal deep learning field.
Different modalities have different feature distributions.
For example, in acoustic modal, tokens of the same sounds tend to have similar features, and for linguistic modal, tokens of the same semantics share similarities accordingly. 
Besides, different modalities have different spatial distributions for the same expression. 
Though acoustic and linguistic modals have consistent in the time dimension, the paired representations are different in length and duration.
These two contradictions should be resolved to make different modalities combined.

In this paper, we propose to employ a CIF aligner with an aligned token similarity loss to solve the problems of multi-modal fusion caused by feature distribution and spatial distributions.

Based on the aligner, the output of the acoustic model is aligned to the same shape as the output of the linguistic model. 
Instead of using ASR, we replace the word embedding layer and even more transformer layers of the linguistic model, with the aligned acoustic output. 
In this way, the combination of two modalities is achieved for SLU tasks.

Our contributions can be summarized as follows:  
(1) We employ the Wave BERT (WaBERT), a novel end-to-end model combining the speech model and the language model to improve the performance of SA tasks.
(2) We modify the CIF mechanism to achieve the monotonic alignment between two different modalities, more specifically, speech and text modalities.
(3) Instead of training from scratch, by employing a pre-trained speech and language model and setting most parameters of WaBERT frozen during training, we achieve short-time and low-resource training.

\section{Related works}
\subsection{Pre-trained speech models}
Recently, self-supervised learning (SSL) has achieved great success in the fields of speech processing.
There are multiple approaches for SSL of speech processing, including wav2vec~\cite{DBLP:journals/corr/abs-1904-05862},
wav2vec 2.0~\cite{DBLP:journals/corr/abs-2006-11477}, HuBERT~\cite{DBLP:journals/corr/abs-2106-07447}, WavLMs~\cite{DBLP:journals/corr/abs-2110-13900},
and data2vec~\cite{DBLP:journals/corr/abs-2202-03555}.

wav2vec~\cite{DBLP:journals/corr/abs-1904-05862} pre-trains a network optimized via a noise contrastive binary classification task.

wav2vec 2.0~\cite{DBLP:journals/corr/abs-2006-11477}, HuBERT~\cite{DBLP:journals/corr/abs-2106-07447}, WavLMs~\cite{DBLP:journals/corr/abs-2110-13900},
and data2vec~\cite{DBLP:journals/corr/abs-2202-03555} all employ BERT-like Masked Acoustic Model(MAM) task for model pre-training, but utilize different features as aligned target labels for prediction.

These methods have shown impressive results for speech processing, especially in phoneme classification and automatic speech recognition (ASR). 
It leverages large amounts of speech data to learn universal speech representations, which can benefit almost all speech downstream tasks by fine-tuning.

However, pre-trained speech models still present some challenges.
On the one hand, in higher-level SLU tasks, satisfying performance is still hard to reach.
Some researches demonstrate that the pre-trained speech models do not learn significant semantic information~\cite{DBLP:journals/corr/abs-2111-10367,https://doi.org/10.48550/arxiv.2203.00648}.
On the other hand, speech data is at a lower abundance and more difficult to obtain compared to text data. 
For example, the text file size of LibriSpeech ASR corpus is 297MB, and wave2vec employs this dataset to achieve SOTA results on ASR tasks.
Meanwhile, for natural language processing (NLP) models, Bert~\cite{DBLP:journals/corr/abs-1810-04805} uses BooksCorpus and English Wikipedia as datasets, which are summed up to 16GB, and Roberta~\cite{DBLP:journals/corr/abs-1907-11692} is trained with datasets Books Corpus, English Wikipedia, CC-News, OpenWebText and Storie, in 161GB, 542 times bigger than LibriSpeech ASR corpus.
The lack of data limits the performance of Speech models.

We propose to overcome these challenges by combining a pre-trained speech model with a pre-trained NLP model.

\subsection{Pre-trained neural language models}
The rapid development in pre-trained neural language models has significantly improved the performance of many NLP tasks.
Models like BERT~\cite{DBLP:journals/corr/abs-1810-04805}, RoBERTa~\cite{DBLP:journals/corr/abs-1907-11692} and DeBERTa~\cite{DBLP:journals/corr/abs-2006-03654} all get acceptable results on NLP tasks.

NLP tasks are similar to SLU tasks but different in data format.
The input data of NLP tasks are texts, while the input data of SLU tasks are audios.
Texts and audios are similar and inter-convertible, and they have enormous common knowledge naturally.
Therefore, NLP models have the potential to be applied in SLU tasks.
However, the different distribution and different lengths between audios and texts prevent NLP models from participating in SLU tasks directly. 
Instead, NLP models are applied in SLU in a more indirect and auxiliary way, the spoken language is recognized as texts by ASR, and then NLP models is fine-tuned for downstream SLU tasks~\cite{DBLP:journals/corr/abs-2111-10367}.
Obviously, this method suffers from errors that occur in the ASR process and loses emotion information by dropping the feature of speech models.

We propose to overcome these challenges by forming an end-to-end model, which combines the pre-trained speech model and the pre-trained NLP model by aligning the outputs of speech and NLP models at word-token level.

\subsection{Forced alignment strategies}
Forced alignment (FA), a strategy to produce the bidirectional mapping between the given text and speech sequences, has been widely used in speech processing tasks for decades.
Conventionally, FA models are based on hidden Markov model
(HMM)~\cite{1165342}, such as Prosodylab-Aligner~\cite{Prosodylab-aligner} and Montreal forced
aligner (MFA)~\cite{mcauliffe2017montreal}.
Also, neural network based FA strategy like NeuFA~\cite{https://arxiv.org/abs/2203.16838} and CIF~\cite{DBLP:journals/corr/abs-1905-11235,https://doi.org/10.48550/arxiv.2203.03582} is proposed.
Even Connectionist Temporal Classification (CTC) can be generally regarded as an FA strategy.

Nevertheless, when applying these methods to align text and speech modalities instead of text and speech sequences, most of the methods become ineffective.
For HMM methods, the hidden states should be discrete, and the number of hidden states should be limited.
Due to words or phonemes being discrete, and the number of words or phonemes being limited,  text can play a role as hidden states in HMM methods. 
For aligning text and speech modals, the tensors that needed to be paired are continuous, and the tensors are unlimited in the account.
Therefore, HMM models can not apply in modalities alignments.
CTC faces the same problem as HMM models.

NeuFA needs ASR and Text-to-Speech (TTS) tasks for joint training, thus, it is not a lightweight construction and not easy for transfer,  which makes it not a  good solution for  modalities alignment

CIF does not require the paired tensor to be discrete or limited in the account, so CIF shows the potential to work in text and speech modals.

We propose to make CIF a better aligner by modifying the aligned token similarity loss.

\begin{figure}
\centering
\includegraphics[width=\textwidth]{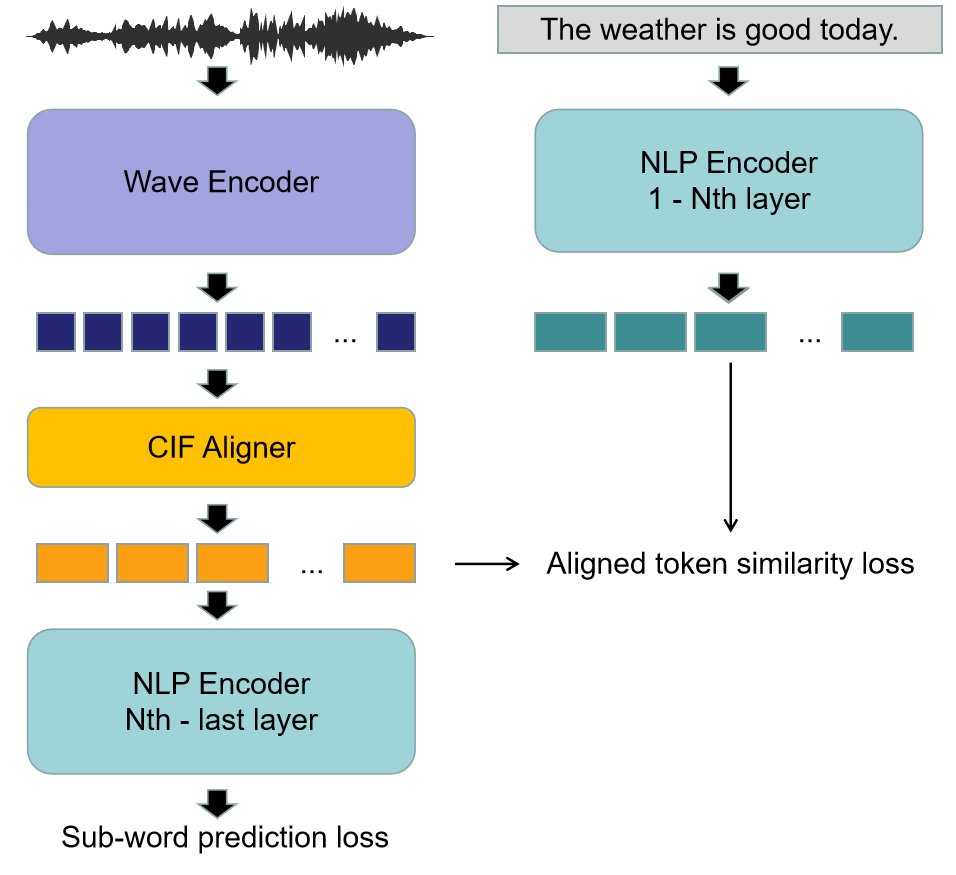}
\caption{WaBERT model overview}
\label{fig:model}
\end{figure}

\section{Approach}

Our model is based on a pre-trained wave encoder and NLP encoder, while any wave and NLP model can be chosen as the component.
Concretely, the recently proposed data2vec and classical BERT are utilized. 
We combine data2vec and BERT to form a new end-to-end model by introducing a CIF aligner.
Figure~\ref{fig:model} shows an overview of our proposed WaBERT model.
We also further improve the alignment between two modals by modifying the align loss. 
More details of this model are provided in the next subsections.

\subsection{Alignment between two modals}
Data2vec extracts acoustic representation vectors from raw audio, and corresponding linguistic representation vectors can be obtained from BERT. 
Suppose the acoustic representation vectors as $A = (a_1, a_2. . . , a_M)$, and the corresponding linguistic representation vectors from $i_{th}$ NLP model layer as
$L_i = (l_1^i, l_2^i,. . . , l_N^i)$,  where $M>N$. 

Due to the acoustic and linguistic representations being consistent in the time dimension, the outputs of the two modalities should be able to be aligned no matter which model layer's output is. 

To avoid collapse, one modality should be frozen during alignment.
For the concern of reducing the information loss of BERT, which contains more information than wave models, the parameters of BERT are fixed.
Inspired by the work\cite{https://doi.org/10.48550/arxiv.2203.03582}, we use the serial CIF~\cite{DBLP:journals/corr/abs-1905-11235} mechanism to achieve the monotonic alignment between the speech and text modalities.

CIF~\cite{DBLP:journals/corr/abs-1905-11235} mechanism is a strategy that by acquiring the length the inputs should be resized to, outputs a latent representation that is consistent with the length, also, a predicted length is provided as another output.
After applying CIF to the output of wav2vec, acoustic representation vectors as $\hat{A} = (\hat{a}_1, \hat{a}_2. . . , \hat{a}_N)$ are obtained.
Regarding $L_i$ as the learning target of the $\hat{A}$, We calculate the loss between $\hat{A}$ and $L_i$.
The work\cite{https://doi.org/10.48550/arxiv.2203.03582} uses the equation below to calculate the loss between two modals for alignment:
\begin{equation} 
cos(x, y) = \frac{x^Ty}{\left \|x\right \|\left \|y\right \|}
\end{equation}
\begin{equation} 
L_{cos} = \sum_{i=0}^N(1 \mbox{-} cos(\hat{a_i}, l_i))
\end{equation}
The similarity between corresponding representations is the only consideration in this loss equation design, and differences between different tokens are ignored.
Emphasizing the difference between different representations would make the adjacent tokens easier to distinguish, and make alignment more precise.
Therefore, we introduce Info Noise Contrastive Estimation (InfoNCE) to calculate the loss between two modalities for alignment.
InfoNCE is defined as below:
\begin{equation} 
InfoNCE(X, Y) =\frac{1}{N} \sum_{i=0}^N(-log(\frac{exp(\frac{cos(x_i,y_i)}{\tau})}{\sum_{j=0}^Nexp(\frac{cos(x_i,y_j)}{\tau})})
\end{equation}
And our aligned token similarity loss for calculating the loss between two modalities for alignment is defined as below:
\begin{equation} he 
L_{InfoNCE} =\frac{1}{2}InfoNCE(\hat{A}, L) +\frac{1}{2}InfoNCE(L, \hat{A})
\end{equation}

The distance between the speech representation lengths that should be aligned to and the predicted length is measured by:
\begin{equation} 
L_{quantity} = N_{predicted} - N
\end{equation}

\subsection{Grafting between two modals}
After aligning the outputs of data2vec with $i_th$ BERT layer, the outputs of data2vec after the CIF aligner, share similar values with the outputs of $i_th$ NLP model.
Grafting between models means replacing a part of the model with another model.
Data2vec can be grafted on BERT by replacing the first to $i_th$ BERT layers with the whole data2vec model, and the new grafted model is supposed to get the same result at the top of the BERT as the input is paired texts. 
Therefore, the original vocabulary ID can be predicted directly with BERT fixed, through this frame, and the BERT-like vocabulary id prediction loss is defined as $L_{subword}$.

\subsection{Training and inference}
The loss $L_total$ for training is defined as:
\begin{equation} 
L_{total} = L_{InfoNCE} + L_{quantity} + L_{subword}
\end{equation}
We employ LibriSpeech ASR corpus for training, and it contains 960 hours of training data.
We train with the adamw optimizer and a batch size of 128, distributed over 8 GPUs. 
The learning rate is linearly ramped up during the first 4000 iterations to its base value of 0.0005.
After this warmup, we decay the learning rate with a linear schedule.
The model is trained for 26 epochs because the loss value is stable enough after 26 epochs.

When coming to inference, the first to $i_{th}$ NLP layer is dropped.
For fine-tuning, with the BERT fixed, the performance of downstream tasks is still acceptable.

\section{Experimental results}
\subsection{CIF alignments}
BERT last layer as labels, data2vec are trained over LibriSpeech ASR corpus, with $L_{subword}$ removed.
 After training of 19 epochs, the loss is stable.
The TIMIT dataset is selected as the test dataset.
The models are evaluated on the MAE and medians of absolute errors of the predicted left and right boundaries at the word level. Then the comparisons are made with CIF with $L_{cos}$ as the baseline.
As shown in Table~\ref{cif-mae}, CIF with $L_{InfoNCE}$ outperforms CIF with $L_{cos}$. 
The MAE is reduced from 426.44 ms to 126.05 ms, and the medians drop from 386 ms to 77 ms, which means that CIF with $L_{InfoNCE}$ always predicts more accurate boundaries than CIF with $L_{cos}$. 
This can also be demonstrated by the accuracies at different tolerances (percentage below a cutoff) evaluated and shown in Table~\ref{cif-acc}.
The accuracies are improved by 0.2825, 0.5532, 0.2893, and 0.0253 at 50, 100 500, and 1000 ms tolerances for word level.

\begin{table}
  \caption{ Performances of the baseline and proposed approaches}
  \label{cif-mae}
  \centering
  \begin{tabular}{lll}
    \toprule
    Approach & mean & median \\
    \midrule
    CIF with $L_{cos}$ & 426.44 & 386 \\
    CIF with $L_{InfoNCE}$ & \textbf{126.05} & \textbf{77} \\
    \bottomrule
  \end{tabular}
\end{table}

\begin{table}
  \caption{ Accuracies at different tolerances for different approaches}
  \label{cif-acc}
  \centering
  \begin{tabular}{lllll}
    \toprule
    Approach & 50 ms & 100 ms & 500 ms & 1000 ms \\
    \midrule
    CIF with $L_{cos}$ & 0.0201 & 0.0559 & 0.6758 & 0.9698 \\
    CIF with $L_{InfoNCE}$ & \textbf{0.3026} & \textbf{0.6091} & \textbf{0.9651} & \textbf{0.9951} \\
    \bottomrule
  \end{tabular}
\end{table}

\begin{figure}
\centering
\begin{minipage}[t]{0.48\textwidth}
\centering
\includegraphics[width=7.5cm]{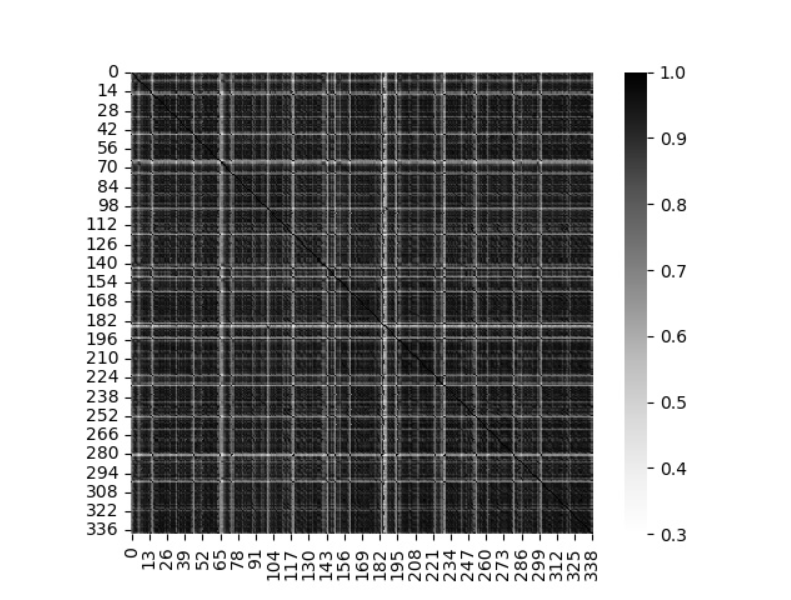}
\caption{The similarity between the acoustic representation after CIF with  $L_{cos}$ and linguistic representation}
\label{fig:cif_sample_a}
\end{minipage}
\begin{minipage}[t]{0.48\textwidth}
\centering
\includegraphics[width=7.5cm]{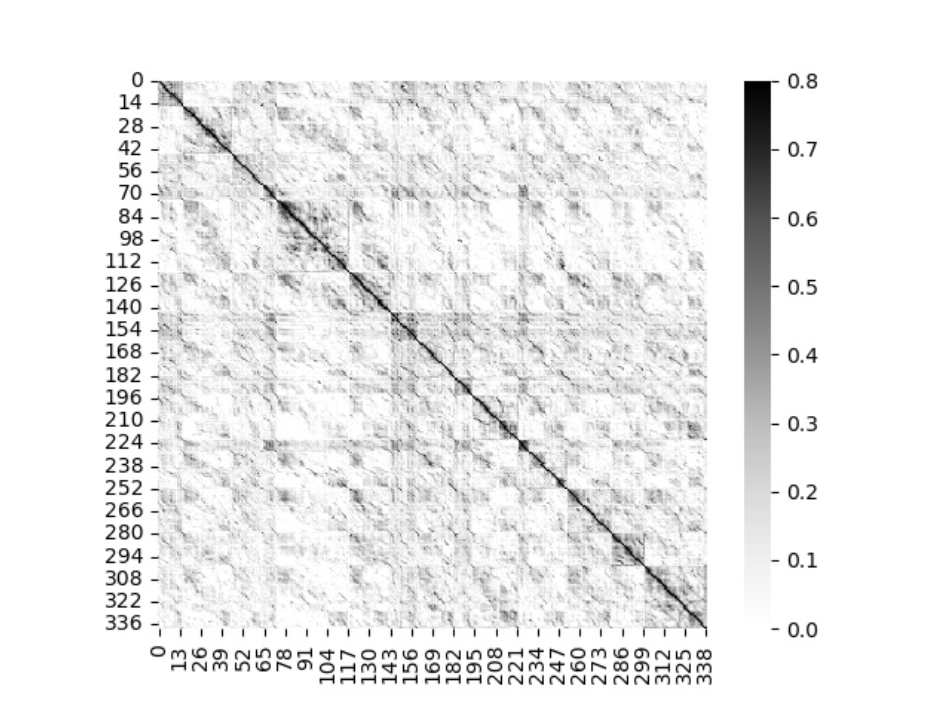}
\caption{The similarity between the acoustic representation after CIF with  $L_{InfoNCE}$ and linguistic representation}
\label{fig:cif_sample_b}
\end{minipage}
\end{figure}

Figure~\ref{fig:cif_sample_a} proves CIF with $L_{InfoNCE}$ is better in other aspects.
It shows the heat-map of the similarity between the acoustic representations  after alignment and the linguistic representations.
And after CIF with $L_{InfoNCE}$, the heat-map shows a valid diagonal, and for CIF with $L_{cos}$, it is still a mess. 

\begin{figure}
\centering
\includegraphics[width=\textwidth]{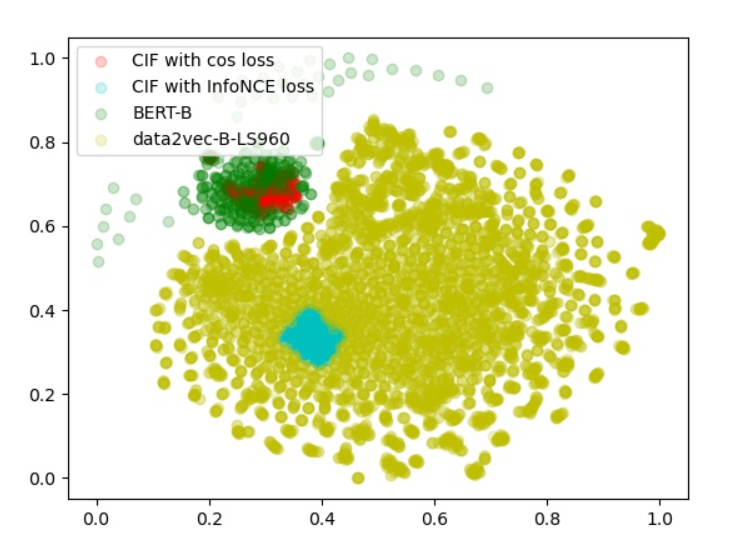}
\caption{The PCA distributions of different outputs}
\label{fig:cif_pca}
\end{figure}

Still, an interesting phenomenon is shown is Figure~\ref{fig:cif_pca}, when using $L_{cos}$, the distribution of acoustic representations is similar to linguistic representations, and drops the initial acoustic distribution.
However, acoustic representations with $L_{InfoNCE}$ remains more initial acoustic distribution, and still make a good performance on alignment.

\subsection{Performances on SLUE SA tasks}
SA refers to classifying a given speech segment as having negative, neutral, or positive sentiment. It is a higher-level task since the semantic content is also very important. For example, negative sentiment can be expressed by disparagement, sarcasm, doubt, suspicion, frustration, etc.~\cite{mohammad2016practical}. We evaluate SA using weighted recall and F1 scores on SLUE benchmark.
The results from Table~\ref{sa-detail} show that, our model outperforms two-stage pipeline approaches and one-stage approaches in dev dataset.

\subsubsection{Comparing to two stage method}
Using the two-stage method, SLU tasks are changed to NLU tasks successfully.
The NLP top-lines and pipeline approaches results in Table~\ref{sa-detail}, show a conclusion:
Errors that occur during ASR, influence the downstream task result. The better the ASR result is the better performance in  SLU tasks. Therefore, None of the methods that use ASR exceed the methods that directly use the ground truth text.

For our end-to-end model, though the wave encoder of our model is not able to give exact features as the ground truth, it provides enough accurate information.
For example, taking our model as the ASR model, a hard example, "never drink any hard liquors such as whisky brandy \textbf{gin or cocktails} with oysters or \textbf{clams} as it is liable to upset you for the rest of \textbf{the evening}" is recognized as "never drink any hard liquors such as whisky brandy \textbf{or or eels} with oysters or \textbf{clamps} as it is liable to upset you for the rest of \textbf{theran}".
Admittedly, there are lots of mistakes in the ASR process, but as shown in Table~\ref{error-analysis}, the right answers all occurred in the top 5 candidates.
Our model outputs features that contain the right information to the NLP model, and avoids the influence of the ASR errors. 
Additionally, the joining of the wave encoder enhances the performance of the whole model, making our result better than the pure BERT model.
Compared to the results of the NLP toplines, our model improves 4.39\% of the recall score and 1.12\% of the F1 score.

\subsubsection{Comparing to one stage method}
Compared to the results of the one-stage method (E2E approaches in Table~\ref{sa-detail}), our model improves 1.19\% of recall score and 0.72\% of F1 score.
The performance proves that the join of an NLP encoder enhances the understanding of spoken language.


\begin{table}
  \caption{ SA detail performance (Recall / F1)}
  \label{sa-detail}
  \centering
  \begin{tabular}{lll|l}
    \toprule
    \multicolumn{3}{c}{Model} & Dev \\
    \cmidrule(r){1-4} 
    Speech model & LM & Text model & Weighted \\
    \midrule
    \multicolumn{3}{l}{\textbf{NLP Toplines :}} \\
    \multirow{ 3}{*}{N/A (GT text)} & \multirow{ 3}{*}{N/A} 
    & BERT-B & 77.8 / 78.9 \\
    \multicolumn{2}{c}{} 
    & DeBERTa-B & 75.5 / 77.3 \\
    \multicolumn{2}{c}{} 
    & DeBERTa-L & 77.7 / 79.3 \\
    \midrule
    \multicolumn{3}{l}{\textbf{Pipeline approaches :}} \\
    W2V2-B-LS960 & - & BERT-B & 77.1 / 78.2 \\
    W2V2-B-LS960 & - & DeBERTa-B  & 74.9 / 76.9 \\
    W2V2-B-LS960 & - & DeBERTa-L & 76.6 / 78.1 \\
    W2V2-L-LL60K & - & DeBERTa-L & 76.5 / 78.0 \\
    \midrule
    W2V2-B-LS960 & $\surd$ & BERT-B & 77.5 / 78.2 \\
    W2V2-B-LS960 & $\surd$ & DeBERTa-B & 73.9 / 75.9 \\
    W2V2-B-LS960 & $\surd$ & DeBERTa-L & 75.9 / 77.7  \\
    W2V2-L-LL60K & $\surd$ & DeBERTa-L & 76.4 / 78.2 \\
    \midrule
    \multicolumn{3}{l}{\textbf{E2E approaches :}} \\
    W2V2-B-LS960 & N/A & N/A & 81.0 / 79.7 \\
    W2V2-B-VP100K & N/A & N/A & 78.8 / 77.0 \\
    HuBERT-B-LS960 & N/A & N/A & 80.5 / 79.4 \\
    W2V2-L-LL60K & N/A & N/A & 79.2 / 79.6 \\
    \midrule
    \multicolumn{3}{l}{\textbf{Ours :}} \\
    data2vec-B-LS960 & N/A & N/A & 80.62/80.28 \\
    data2vec-B-LS960 & N/A & BERT-B(w/o 1-3 layers) & \textbf{82.19}/\textbf{80.42} \\
    data2vec-B-LS960 & N/A & BERT-B(w/o 1-6 layers)  & 80.42/79.92 \\
    data2vec-B-LS960 & N/A & BERT-B(w/o 1-9 layers)  & 80.83/80.37 \\
    data2vec-B-LS960 & N/A & BERT-B(w/o 1-12 layers) & 80.73/80.40 \\
    \bottomrule
  \end{tabular}
\end{table}

\begin{table}
  \caption{Error analysis of sub-word prediction sample}
  \label{error-analysis}
  \centering
  \begin{tabular}{l|lll|llll|ll}
    \toprule
    \textbf{Labels} & gin & or & cocktails & \multicolumn{4}{c}{clams} & the & evening \\
    \midrule
    \textbf{Top-1} & or & \textbf{or} & eels & \textbf{cl} & -amp & \textbf{-s} & as & \textbf{the} & -ran \\
     \textbf{Top-2} & \textbf{gin} & \textbf{cock} & \textbf{tails} & cr & \textbf{-am} & \textbf{-ams} & it & , & \textbf{evening} \\
     \textbf{Top-3} & gill & char & beans & or & -umb & -mes & \textbf{-s} & . & he \\ 
     \textbf{Top-4} & chin & a & turtles & the & -ab & as & that & a & ang \\
     \textbf{Top-5} & and & og & ears & br & -lar & -as & in &to & night \\
    \bottomrule
  \end{tabular}
\end{table}

\section{Ablation studies}
Table~\ref{sa-detail} shows that, when aligning data2vec's outputs to the outputs of $3_{rd}$, $6_{th}$, $9_{th}$ and $12_{th}$ BERT layer, the result of $3_{rd}$ and $12_{th}$ layer is better than others. The output of $3_{rd}$ is more near to word embedding layers, and $12_{th}$ is more close to the classification layer, which are under more strict constraints than other compared layers, thus, the outputs are more independence and easy to separate. align technique is easier to show its effort.

For the comparison of output of $3_{rd}$ and $12_{th}$ layers, the $3_{rd}$ layer method keeps more BERT structure, and this part gives model the abilities to outperform the $12_{th}$ layers method.

\section{Conclusions}
Multi-modal alignment is different, the labels are more flexible and dependent compared to label-fixed FA. 
This paper showed that CIF can be used as a multi-modal aligner when a proper align loss is chosen. The main reason is that this strategy convergences feature distributions and space distributions.

By using WaBERT, we were able to integrate audio-specific information and language knowledge to improve the performance of SA tasks in a short-time and low-resource training process.

\section{Further studies}
(1) The loss between sub-word prediction and the label is calculated for all tokens, though some researches show that only calculating the loss between labels and masked tokens leads to better performance. This method would be applied in our further studies.

(2) In this paper, we only adopt BERT as the NLP encoder and data2vec as the wave encoder, the influence of different combinations of models should be another direction of further studies.

(3) Most of the WaBERT parameters are frozen during training. However, making all parameters trainable might be a way to improve the performance. We will try to do speech pre-train tasks, NLP pre-train tasks, and our aligned tasks together for joint training in the next step.

(4) The aligner only aligns one layer in our method, in the next step, we will align different layers respectively at the same time, to achieve a more precise alignment.



\bibliographystyle{splncs04}
\bibliography{ref}

\begin{thebibliography}{10}
\providecommand{\url}[1]{\texttt{#1}}
\providecommand{\urlprefix}{URL }
\providecommand{\doi}[1]{https://doi.org/#1}

\bibitem{DBLP:journals/corr/abs-2202-03555}
Baevski, A., Hsu, W., Xu, Q., Babu, A., Gu, J., Auli, M.: data2vec: {A} general
  framework for self-supervised learning in speech, vision and language. CoRR
  \textbf{abs/2202.03555} (2022), \url{https://arxiv.org/abs/2202.03555}

\bibitem{DBLP:journals/corr/abs-2006-11477}
Baevski, A., Zhou, H., Mohamed, A., Auli, M.: wav2vec 2.0: {A} framework for
  self-supervised learning of speech representations. CoRR
  \textbf{abs/2006.11477} (2020), \url{https://arxiv.org/abs/2006.11477}

\bibitem{DBLP:journals/corr/abs-2110-13900}
Chen, S., Wang, C., Chen, Z., Wu, Y., Liu, S., Chen, Z., Li, J., Kanda, N.,
  Yoshioka, T., Xiao, X., Wu, J., Zhou, L., Ren, S., Qian, Y., Qian, Y., Wu,
  J., Zeng, M., Wei, F.: Wavlm: Large-scale self-supervised pre-training for
  full stack speech processing. CoRR  \textbf{abs/2110.13900} (2021),
  \url{https://arxiv.org/abs/2110.13900}

\bibitem{https://doi.org/10.48550/arxiv.2203.03582}
Deng, K., Cao, S., Zhang, Y., Ma, L., Cheng, G., Xu, J., Zhang, P.: Improving
  ctc-based speech recognition via knowledge transferring from pre-trained
  language models (2022). \doi{10.48550/ARXIV.2203.03582},
  \url{https://arxiv.org/abs/2203.03582}

\bibitem{DBLP:journals/corr/abs-1810-04805}
Devlin, J., Chang, M., Lee, K., Toutanova, K.: {BERT:} pre-training of deep
  bidirectional transformers for language understanding. CoRR
  \textbf{abs/1810.04805} (2018), \url{http://arxiv.org/abs/1810.04805}

\bibitem{DBLP:journals/corr/abs-1905-11235}
Dong, L., Xu, B.: {CIF:} continuous integrate-and-fire for end-to-end speech
  recognition. CoRR  \textbf{abs/1905.11235} (2019),
  \url{http://arxiv.org/abs/1905.11235}

\bibitem{Prosodylab-aligner}
Gorman, K., Howell, J., Wagner, M.: Prosodylab-aligner: A tool for forced
  alignment of laboratory speech. vol.~39 (09 2011)

\bibitem{DBLP:journals/corr/abs-2006-03654}
He, P., Liu, X., Gao, J., Chen, W.: Deberta: Decoding-enhanced {BERT} with
  disentangled attention. CoRR  \textbf{abs/2006.03654} (2020),
  \url{https://arxiv.org/abs/2006.03654}

\bibitem{DBLP:journals/corr/abs-2106-07447}
Hsu, W., Bolte, B., Tsai, Y.H., Lakhotia, K., Salakhutdinov, R., Mohamed, A.:
  Hubert: Self-supervised speech representation learning by masked prediction
  of hidden units. CoRR  \textbf{abs/2106.07447} (2021),
  \url{https://arxiv.org/abs/2106.07447}

\bibitem{DBLP:journals/corr/abs-1907-11692}
Liu, Y., Ott, M., Goyal, N., Du, J., Joshi, M., Chen, D., Levy, O., Lewis, M.,
  Zettlemoyer, L., Stoyanov, V.: Roberta: {A} robustly optimized {BERT}
  pretraining approach. CoRR  \textbf{abs/1907.11692} (2019),
  \url{http://arxiv.org/abs/1907.11692}

\bibitem{mcauliffe2017montreal}
McAuliffe, M., Socolof, M., Mihuc, S., Wagner, M., Sonderegger, M.: Montreal
  forced aligner: Trainable text-speech alignment using kaldi. In: Interspeech.
  vol.~2017, pp. 498--502 (2017)

\bibitem{mohammad2016practical}
Mohammad, S.: A practical guide to sentiment annotation: Challenges and
  solutions. In: Proceedings of the 7th workshop on computational approaches to
  subjectivity, sentiment and social media analysis. pp. 174--179 (2016)

\bibitem{1165342}
Rabiner, L., Juang, B.: An introduction to hidden markov models. IEEE ASSP
  Magazine  \textbf{3}(1),  4--16 (1986). \doi{10.1109/MASSP.1986.1165342}

\bibitem{https://doi.org/10.48550/arxiv.2203.00648}
Sanabria, R., Hsu, W.N., Baevski, A., Auli, M.: Measuring the impact of
  individual domain factors in self-supervised pre-training (2022).
  \doi{10.48550/ARXIV.2203.00648}, \url{https://arxiv.org/abs/2203.00648}

\bibitem{DBLP:journals/corr/abs-1904-05862}
Schneider, S., Baevski, A., Collobert, R., Auli, M.: wav2vec: Unsupervised
  pre-training for speech recognition. CoRR  \textbf{abs/1904.05862} (2019),
  \url{http://arxiv.org/abs/1904.05862}

\bibitem{DBLP:journals/corr/abs-2111-10367}
Shon, S., Pasad, A., Wu, F., Brusco, P., Artzi, Y., Livescu, K., Han, K.J.:
  {SLUE:} new benchmark tasks for spoken language understanding evaluation on
  natural speech. CoRR  \textbf{abs/2111.10367} (2021),
  \url{https://arxiv.org/abs/2111.10367}

\end{thebibliography}

\end{document}